\documentclass[11pt]{article}

\usepackage[final]{acl}

\usepackage{times}
\usepackage{latexsym}

\usepackage[T1]{fontenc}

\usepackage[utf8]{inputenc}

\usepackage{microtype}

\usepackage{inconsolata}

\usepackage{graphicx}

\usepackage{booktabs}
\usepackage[most]{tcolorbox}

\newtcolorbox{promptbox}[1]{
    colback=gray!5,       
    colframe=black!75,    
    fonttitle=\bfseries,
    title=#1,             
    sharp corners,        
    enhanced,
    width=\textwidth,     
    boxrule=0.5pt,
}
\usepackage{float}

\title{Towards Just-in-Time Adaptive Feedback:\\Enhancing Student Learning via Knowledge-Grounded LLM}

\author{Younghun Lee$^{\dagger}$, Amir Bralin$^{\ddagger}$, Nobel Sanjay Rebello$^{\ddagger}$$^{\S}$, Dan Goldwasser$^{\dagger}$\\
       $^{\dagger}$Department of Computer Science\\
       $^{\ddagger}$Department of Physics and Astronomy\\
       $^{\S}$College of Education\\
       Purdue University \\
        \texttt{\{younghun,abralin,rebellos,dgoldwas\}@purdue.edu}}

\begin{document}
\maketitle
\begin{abstract}
Educational interventions are effective tools for enhancing student learning. While Large Language Models (LLMs) allow for generating adaptive feedback at scale, current studies lack clear methodologies for providing Just-in-Time (JiT) feedback in authentic instructional settings. In this paper, we present a framework that provides adaptive feedback by grounding LLMs with domain-specific expert knowledge. Our approach collects written reasoning logic (strategy essays) from students, analyzes potential error types based on the content of that reasoning, and delivers non-intrusive feedback designed to clarify missing or incorrect concepts. We deploy this framework in a large-scale university course ($N > 1,000$), where it improved student performance by over 80\% compared to previous semesters. Lastly, we validate the framework’s pedagogical utility by analyzing the learning trajectories; we demonstrate how iterative conversations with LLM facilitate shifting one's misconception to correct understanding.
\end{abstract}

\begin{figure*}[t]
    \centering
    \includegraphics[width=\textwidth]{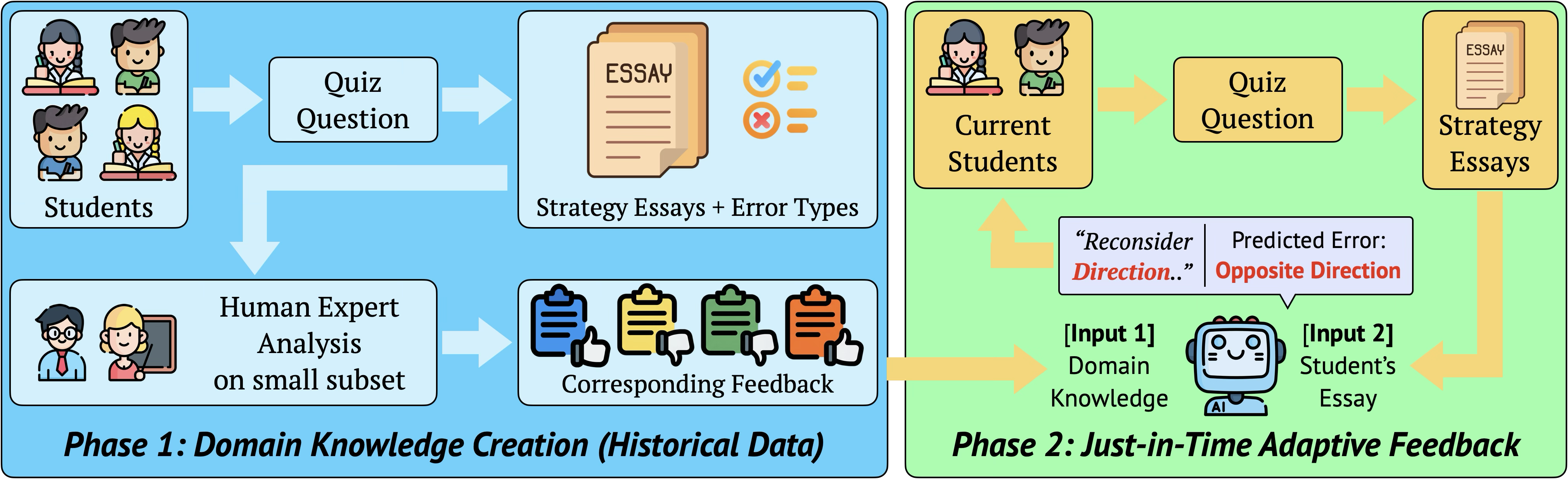}
    \caption{    \textbf{Overall framework} of our Just-in-Time adaptive feedback LLM. In the first phase, we obtain domain knowledge from human experts regarding the appropriate feedback for students' strategy essays and the types of errors they made. At deployment, we ground the LLM with this knowledge and provide adaptive feedback to the students based on their written strategy essays.}
    \label{fig:intro-figure}
\end{figure*}

\section{Introduction}

In STEM education, students are often observed to engage in ``Recursive Plug-and-Chug''---also known as ``Formula Hunting''---where the goal is simply to fill a slot in a formula rather than to make sense of the problem \citep{chi1981categorization, tuminaro2007elements}. Existing studies indicate that while experts analyze problems based on their deep logical structure and principles, novices are often primed to surface features, leading to unsuccessful problem-solving approaches. \citep{mestre1993promoting}. 

Strategy writing has emerged as a powerful intervention to tackle this problem. Research has shown that asking students to articulate their strategies for solving problems can improve their problem-solving skills \citep{leonard1996using}. With a combination of argumentation and appropriate prompts, students focus more on deep structure rather than the surface features of the problem and starts solving the problem with conceptual analysis, avoiding novice, unproductive strategies \citep{dufresne1992constraining, mestre1993promoting,docktor2010conceptual,rebello2019using}.

Despite the benefits, strategy writing is difficult to implement at scale. The primary impediment is the feedback bottleneck. Feedback is most effective when it is integrated into the learning process through formative assessment \citep{hattie2007power}, and provided prior to completion \citep{henderson2021usefulness}. However, providing real-time, formative feedback to thousands of students is prohibitively time-consuming for human instructors. Thus, the true potential of strategy writing to enhance student learning remains largely unrealized. The advance of generative Large Language Models (LLMs) offers a promising pathway toward overcoming the scalability barrier. Recent work proposes adaptive scaffolding frameworks that utilizes LLMs to help students understand course materials \citep{taneja2024jill, kweon2025large, kestin2025ai, zhang2025simulating}. However, they lack exploration of intervention-oriented interactions as well as grounding the models with domain-specific knowledge.

In this paper, we address these challenges by exploring dimensions of adaptive feedback and implementing a framework that provides Just-in-Time interventions using knowledge-grounded LLMs. First, we collect strategy essays from students over a couple of semesters and obtain historical data that connect the essays to the types of errors students had made. Human experts then annotate feedback on a small portion of the data considering the content of the essays, as well as the error types. Using these annotated instances as few-shot examples, the LLM identifies misconceptions in the strategy essays and provide real-time, adaptive feedback. Rather than providing a `shortcut' to the correct answer, the feedback directs the student's attention to the underlying concepts and their reasoning logic (see Figure \ref{fig:feedback-anno} and Figure \ref{fig:conv-sample} for examples). This ensures the intervention disrupts the `formula-hunting' heuristic by letting students reflect on the problem's deep structure. Figure \ref{fig:intro-figure} illustrates the overall framework we propose.

We experiment our framework on a large-scale university physics course ($N > 1,000$). Experimental results show that the overall student performance is improved when the feedback framework is deployed, by over 80\% compared to previous semesters. Based on further analyses on self-reported survey and conversational instances, we argue that the framework effectively facilitates a shift from initial misconceptions to the correct understanding.


\section{Related Work}
Educational research has shown that feedback is one of the important drivers of learning \cite{wisniewski2020power, foster2024impact}. Feedback can facilitate improvements in learners’ understanding and skills \cite{henderson2021usefulness} by narrowing the gap between actual and desired performance \cite{burgess2020feedback}. The effectiveness of feedback increases with the information that it contains. Previous research \cite{kluger1996effects} has also shown that moderators such as timing, specificity, and task complexity affect how learners receive and use feedback \cite{hattie2018visible, brooks2019matrix}.

Feedback is effective if it is sufficiently detailed \cite{price2010feedback}, usable \cite{winstone2017d}, and facilitates change \cite{ryan2016written}, such that learners can test their new understandings \cite{pitt2017now}. In asynchronous and isolated online settings \cite{orlando2016comparison}, interactive dialogues can be especially useful \cite{wolsey2008efficacy} as students cannot easily interact with their peers \cite{furnborough2009adult} which put significant weight on the feedback comments they receive \cite{ortiz2005college}.


Recent research on educational LLM applications has shifted from general assistance to specialized interventions. For instance, \citet{phung2024automating} used GPT-4 as a teacher to provide non-intrusive hints but relied on LLM-simulated student agents using a weaker LLM (GPT-3.5), rather than applying it to real classroom settings. In contrast, \citet{dai2023can} and \citet{jia2024llm} designed LLM feedback to apply to students, but their evaluation metrics focused on student-reported helpfulness survey and qualitative analysis of feedback content, rather than measuring actual improvements in student performance. More recent studies have shown LLM-generated feedback can help improve student performance in class. \citet{Hashmi2025} used an LLM-based Socratic chatbot to scaffold expert-like reasoning. \citet{zhang2025simulating} implemented a persona-driven multi-agent dialogue system, showing the student's performance in problem-solving improves as they interact more with LLMs. \citet{kestin2025ai} integrated an AI tutor in a college Physics course and compared student's performance on a quiz between an AI tutor and in-class lessons. The results show that when the model is applied to a class size of 200, the AI tutor almost doubled the learning gains compared to in-class lessons.

In this paper, we introduce three novel contributions. Unlike reactive chatbots that are susceptible to formula-hunting behaviors of the students, our system proactively analyzes the reasoning logic within the strategy essays that guides students to focus on deep structures in solving quiz problems. Additionally, we ground LLMs with domain-specific knowledge that is carefully annotated by human experts, which brings significant benefits compared to persona-based prompt engineering. Lastly, we provide a robust, automated intervention that remains pedagogically sound at a massive scale ($N > 1,000$).

\section{Problem Formulation}
\subsection{Instructional Setting}

Our research context is a large enrollment calculus-based physics course for engineers and physical scientists at a large U.S. Midwestern land grant university. The focus of the course is the development of problem-solving skills focused on applying key physics principles pertaining to mechanics across three units: Newton’s Laws and linear momentum, work and energy, and angular momentum. The annual enrollment of the course is around 3,300 students (1,500 in fall, 1,800 in spring), about 25\% women, 10\% underrepresented minorities, and 8\% international students. 

Each week, students complete an online quiz administered via the Learning Management System, along with an online proctoring system to maintain integrity. Students have 40 minutes to complete each online quiz. In this work, we focus on the quiz data from four consecutive semesters, from the Fall 2024 to the Spring 2026 semester.

\subsection{Strategy Essays}
\label{sec:strategy-essay}
Strategy essays refer to the student's written reasoning logic in solving the quiz problem \cite{leonard1996using}. For one of the quiz questions, students are instructed to write a strategy essay that is at least 50 words long and does not include any numbers, symbols, or formulae. This was designed to evaluate the consistency between a student's written reasoning logic and their performance on the quizzes. To maintain the quality of the essays, the instructors offer extra credit\footnote{Extra credits equal to 10\% of the total points on the quiz} for writing a strategy essay. In the Fall 2024 semester, for example, we gathered 11,948 essays from 1,418 students from a total of 11 quizzes.


\subsection{Desired Properties of LLM Feedback}
\label{sec:terminology-framework}
LLM feedback refers to the LLM-generated text that is given to each student. The primary purpose of LLM feedback is to help students better understand the course materials related to the quiz problem and solve it correctly. 

One of the most critical constraints in providing feedback is \textbf{to facilitate the student's own problem-solving process}. The LLM should not disclose a direct solution or correct answers to the quiz problems when generating feedback to students.
Another aspect we consider in designing the feedback is \textbf{adaptiveness}; rather than generating universal feedback, it is conditioned on student-specific features such as their in-class performance (e.g. pre-semester assessment scores, midterm exam scores, other quiz scores, etc.) and their strategy essays. 
Lastly, we focus on \textbf{grounding LLMs with domain-specific knowledge}. General-purpose LLMs often exhibit unreliable performance in understanding college-level STEM courses that require multi-step logic \citep{arora2023have,pang2025physics}. To minimize the knowledge gap, we distill domain-specific knowledge from experts (i.e. instructors) and ground LLM generations with it. 

\subsection{Post-Hoc vs. Just-in-Time (JiT)}
LLM feedback can be presented to students in two different ways: after students finish solving the quiz (post-hoc) and in real-time while they solve it (JiT). These two methods serve different benefits. In this paper, we utilize LLM post-hoc feedback to mainly survey the student's preference for the feedback style. Students' preference patterns ultimately help identify how LLM feedback needs to be generated. We consider LLM JiT feedback as a means of real-time interventions which help students solve the quiz problem.


\section{Designing Adaptive Feedback}

\begin{figure}[t]
    \centering
    \includegraphics[width=\columnwidth]{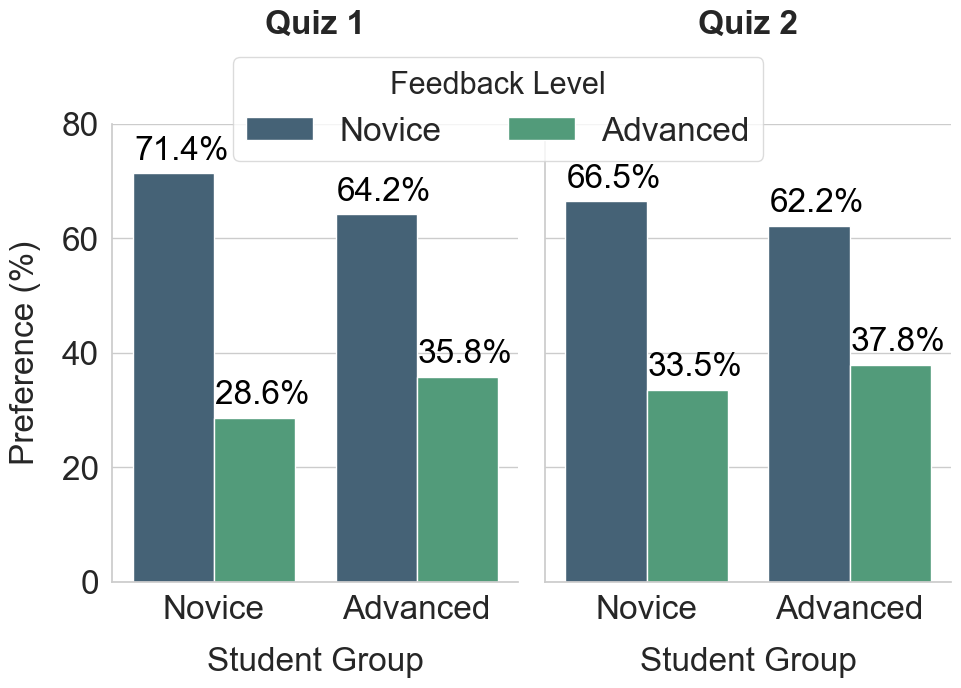}
    \caption{\textbf{Survey results regarding preference} between novice and advanced feedback in two quizzes. Regardless of their level of knowledge, students prefer LLM feedback that is targeted to the novice group.}
    \label{fig:survey-results}
\end{figure}

\subsection{Student Preference Survey}
Existing studies suggest the necessity of adaptive guidance that aligns with the student's level of knowledge; unlike the way novice students perform better with detailed explanations, more experienced learners find it redundant and even show negative consequences \citep{kalyuga2007expertise,walker2012noticing,albacete2019impact}. Since there is limited work that verifies this finding with LLM-based feedback, we conduct a series of surveys to find out whether students with different knowledge levels show a contrasting preference for the complexity of LLM feedback.

We provide post-hoc LLM feedback to the students for two quizzes in the Fall 2025 semester. Students are asked to write a short strategy essay that they employ to solve the problems. After having strategy essays from the students, we prompt LLMs to generate two different versions of the feedback, one for the novice-level and the other for the advanced-level. LLMs are instructed to differentiate not only linguistic styles such as tone and vocabulary, but also how feedback is framed with respect to the focus and goal\footnote{We used GPT-5.1 to generate post-hoc feedback. Detailed settings for prompting are described in Appendix \ref{sec:appendix-preference-survey}}. Students are instructed to participate in the survey on a separate webpage showing two feedback texts, one targeting the novice-level and the other for the advanced-level, and are asked to indicate which version they prefer. The presentation order of the feedback was randomized to mitigate potential order effects and minimize response bias. Figure \ref{fig:survey-webapp} shows the survey webpage (see Appendix \ref{sec:appendix-preference-survey}).

To find the relevance between the student's level of knowledge and the complexity of the LLM feedback, we categorize students into two buckets, novice and advanced. Students were categorized as novice if they performed below average for all relevant assessments; this includes a pre-semester assessment\footnote{Energy and Momentum Conceptual Survey \cite{singh2016multiple}}, mid-term exam, and relevant previous quizzes. In contrast, students who consistently exceeded the average score for all relevant assessments were categorized as advanced.

Figure \ref{fig:survey-results} shows the results. Although the portion of students who favor advanced-level feedback is larger among students with advanced knowledge, both novice and advanced groups prefer feedback that is targeted to the novice group. This implies that LLMs are inadequate to generate advanced feedback that is pedagogically effective, suggesting that simply instructing LLMs to increase vocabulary complexity and reduce scaffolding may hinder rather than help the learning process.
\begin{figure}[t]
    \centering
    \includegraphics[width=0.9\columnwidth]{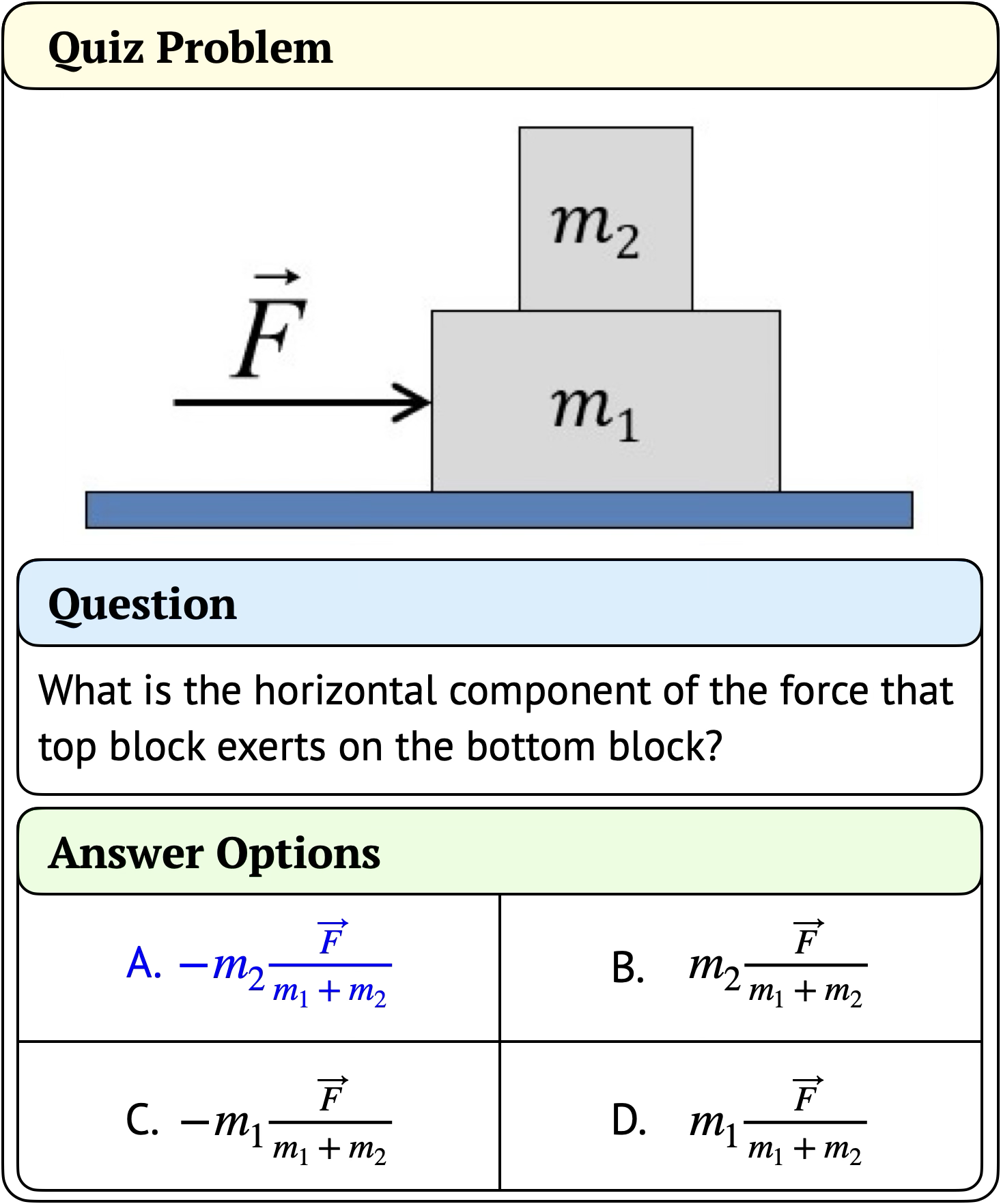}
    \caption{\textbf{Example quiz problem.} While A is correct, other options map to common errors: B represents a \texttt{direction} error (misunderstanding force directions); C represents a \texttt{position} error (incorrect mass/location consideration); and D represents confusion involving both concepts ( \texttt{position-direction} error).}
    \label{fig:quiz3}
\end{figure}
\subsection{Just-in-Time Intervention}
Using insights from the survey results, we focus on a different aspect of adaptive LLM feedback for JiT intervention: \textbf{error types}. 

All incorrect multiple-choice options in the quizzes are designed to elicit specific error types. Consider an example quiz in Figure \ref{fig:quiz3}. The question asks to find the force that the top block exerts on the bottom block, given a certain amount of horizontal force, $\vec{F}$, is applied to the bottom block and the two blocks move together. The correct answer is multiplying the mass of the top block by acceleration which is the opposite direction to the $\vec{F}$. Common errors can be made by not considering that the direction has to be opposite (B. \texttt{direction} error), or by multiplying the mass of another object (C. \texttt{position} error), or by having confusion involving both concepts (D. \texttt{position-direction} error). We posit that LLMs offer effective interventions by predicting a student's likely error type and providing adaptive feedback that guides them to reconsider those aspects in their final answer.
\begin{figure}[t]
    \centering
    \includegraphics[width=0.9\columnwidth]{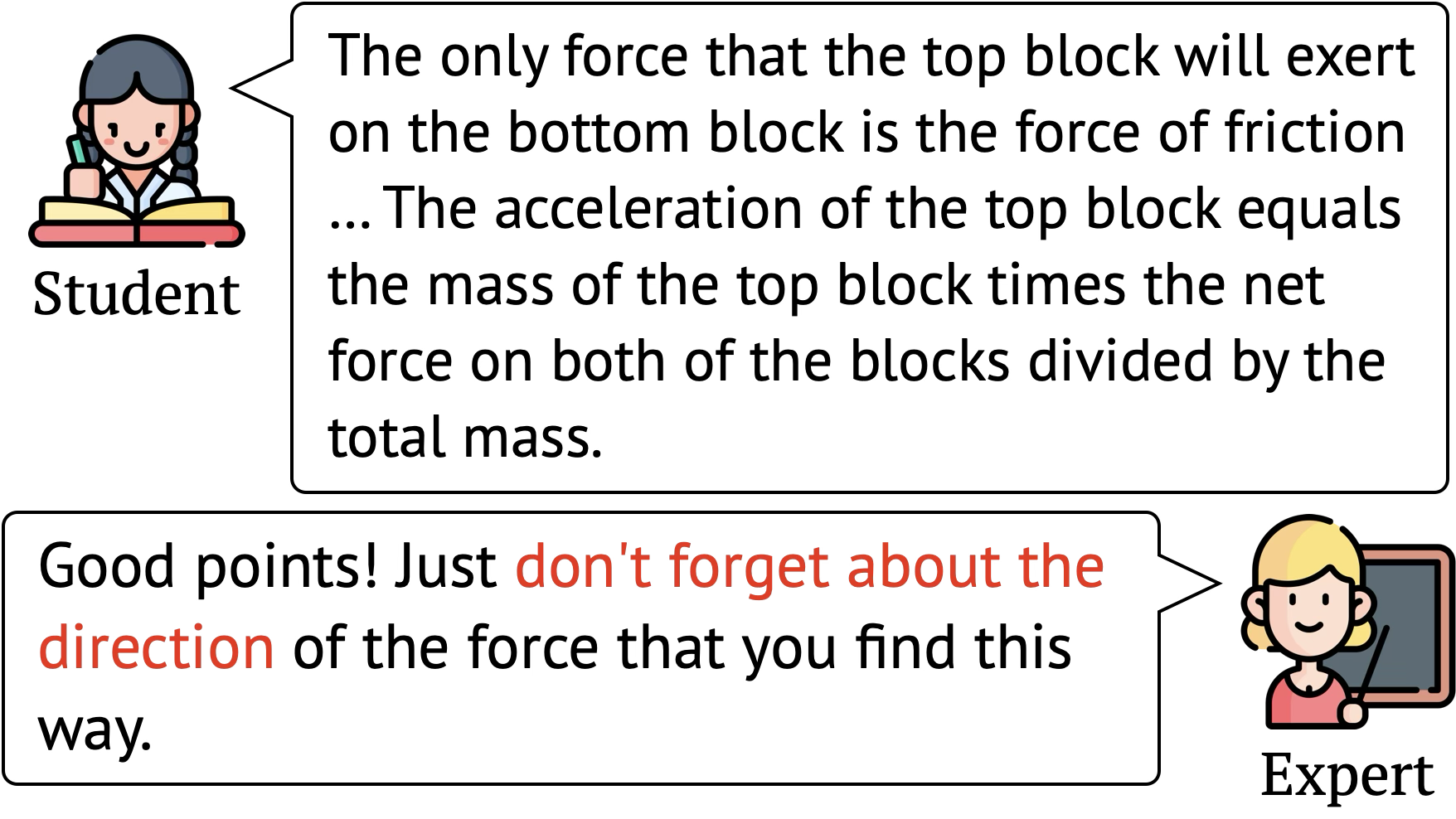}
    \caption{Example of a strategy essay and annotated feedback. Although the student in this example got the correct answer, the feedback mentions the importance of considering direction as it was missing in the essay.}
    \label{fig:feedback-anno}
\end{figure}

\subsection{Grounding LLM Feedback with Domain Knowledge}
As stated in \ref{sec:terminology-framework}, LLMs need to be grounded with domain-specific knowledge to provide more effective feedback with a better understanding of the concepts. We integrate human-expert knowledge into LLMs' feedback generation process. 

Using historical data from previous semesters, we collected students' strategy essays and their corresponding answer choices, enabling us to associate written reasoning with particular error types. Teaching experts (i.e. faculty and instructors) then manually wrote feedback tailored to the content of these essays and error types. 

Such nuanced feedback annotation is naturally a challenging task for LLMs. For example, students who selected the correct answer might still write imperfect strategy essays by not explicitly mentioning the force direction. Conversely, a semantically similar essay written by another student might lead to the \texttt{direction} error. Human experts are more competent in capturing these subtle differences between strategy essays and answer choices. Figure \ref{fig:feedback-anno} illustrates an example of the annotation process, where experts identify missing concepts regardless of whether the student's final answer was correct.

Once the feedback annotation is completed for a small representative set (50 examples), they serve as the few-shot examples for in-context learning, enabling the real-time classification of error types using student's strategy essays. Recognizing that strategy essays are often non-trivial to categorize into potential error types, we move beyond single-label prediction; we prompt the LLMs to identify both the primary and secondary most likely labels. The model then produces feedback via Chain-of-Thought reasoning \citep{wei2022chain}, conditioned on this expanded label space to ensure more robust instructional support. Detailed explanations regarding prompts and JiT intervention framework are provided in Appendix \ref{sec:appendix-live-feedback}.

\section{Experiments and Results}

\subsection{Error Type Predictions}
\begin{table}[t]
\centering
\footnotesize
\begin{tabular}{@{}lcc@{}}
\toprule
\textbf{Method} & \textbf{Accuracy} & \textbf{Macro F1} \\ \midrule
Fine-tuned Classifier & 50.00$^{\pm \text{0.001}}$ & 28.39$^{\pm \text{0.001}}$ \\
Zero-shot LLM & 21.00$^{\pm \text{0.004}}$ & 12.74$^{\pm \text{0.006}}$ \\
Zero-shot LLM w/ Secondary & 37.00$^{\pm \text{0.003}}$ & 28.34$^{\pm \text{0.005}}$ \\
Few-shot LLM & 42.42$^{\pm \text{0.001}}$ & 33.50$^{\pm \text{0.002}}$ \\
Few-shot LLM w/ Secondary & \textbf{60.61}$^{\pm \text{0.002}}$ & \textbf{54.24}$^{\pm \text{0.003}}$ \\ \bottomrule
\end{tabular}
\caption{Classification performance of different models in predicting error types (i.e. correct, direction, position, position-direction) given strategy essay. LLMs perform better when they are prompted with few-shot examples and Chain-of-Thought reasoning while considering secondary label.}
\label{tab:model_results}
\end{table}

In this experiment, we measure the accuracy of error type prediction across different models. The task is formulated as a multi-class classification where the model gets a strategy essay as input and predicts one of four classes: \texttt{correct}, or one of the three error types---\texttt{direction}, \texttt{position}, and \texttt{position-direction}. Table \ref{tab:model_results} describes the accuracy and macro F1 score of the models.

As a baseline, we first implement a sentence classifier using a pre-trained BERT-based checkpoint using the quiz data from Fall 2024 semester. Since the number of data instances is small (1.1K), the trained model shows zero accuracy on labels with insufficient amount of training instances, results in a very low macro F1 score.

\begin{figure*}[t]
    \centering
    \includegraphics[width=0.9\textwidth]{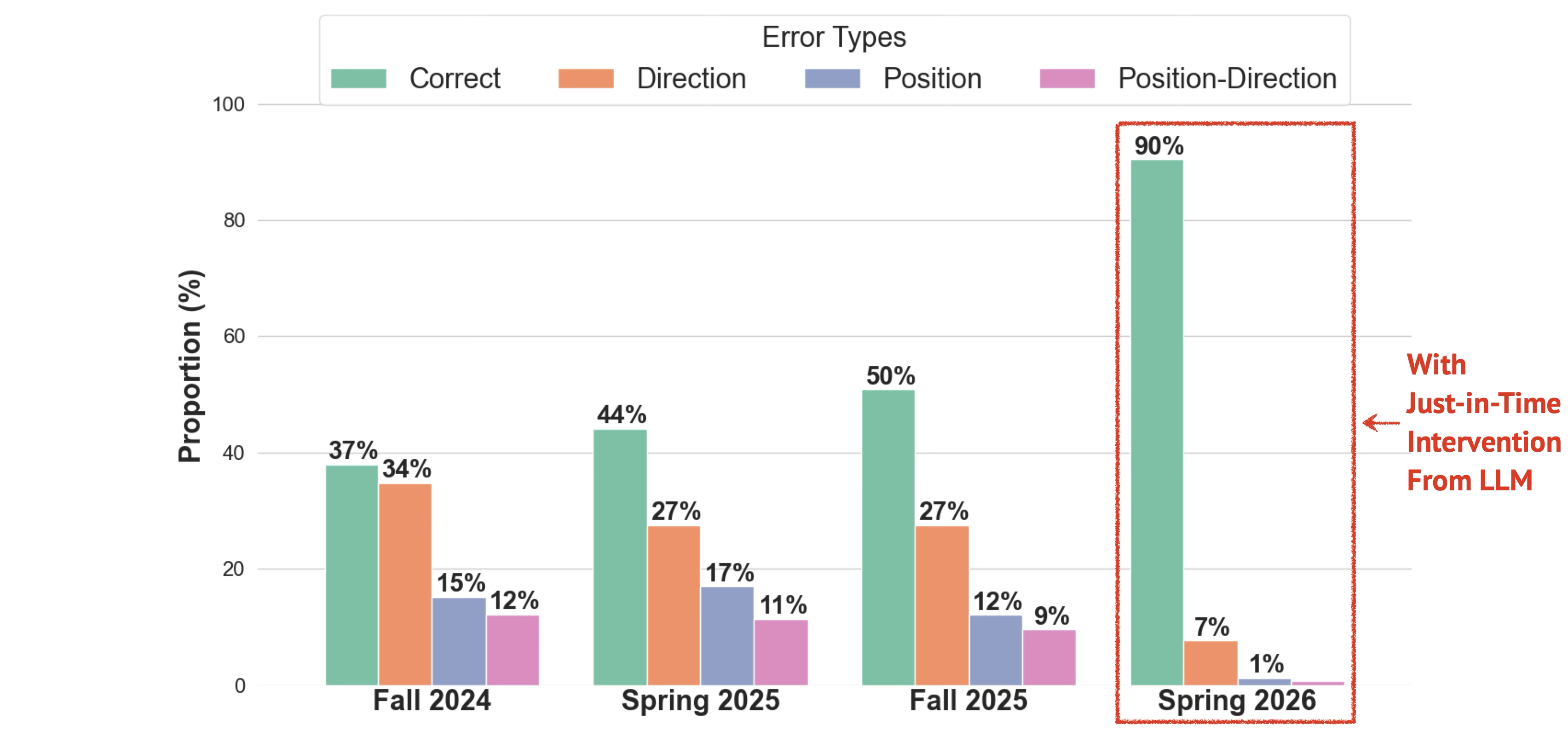}
    \caption{Student's performance in the same quiz question over four semesters. When the students solved the problem with the LLMs providing JiT adaptive intervention, the overall portion of correct students boosted significantly.}
    \label{fig:performance-improvement}
\end{figure*}
As the number of training data instances is limited, we use high-performance LLMs to classify the error types without fine-tuning. We try prompting the LLMs with zero-shot and few-shot settings (3 examples per label). Although adding few-shot examples improve the performance, the accuracy and macro F1 scores are not strong enough. After we prompt the model to consider the secondary most likely labels, the performance increases significantly, implying that the LLM can generate more robust feedback with expanded label space. Detailed explanations regarding prediction models are described in Appendix \ref{sec:appendix-model-accuracy}.

\subsection{Quiz with JiT Interventions}
We administered JiT interventions during a specific quiz in the Spring 2026 semester. The problem and its corresponding multiple-choice options are illustrated in Figure \ref{fig:quiz3}. Students accessed the quiz through their standard Learning Management System, with the LLM feedback interface integrated directly into the problem description. A comprehensive view of the user interface and the interaction flow is provided in Figure \ref{fig:feedback-prompt} (see Appendix \ref{sec:appendix-live-feedback}). Upon engaging with the system, students were informed that the LLM would not provide direct solutions or answers. Instead, they were instructed to use the framework to get advice on specific concepts requiring further attention, based on the content of their strategy essays.

\subsection{Student Performance Improvements}

\begin{figure}[t]
    \centering
    \includegraphics[width=\columnwidth]{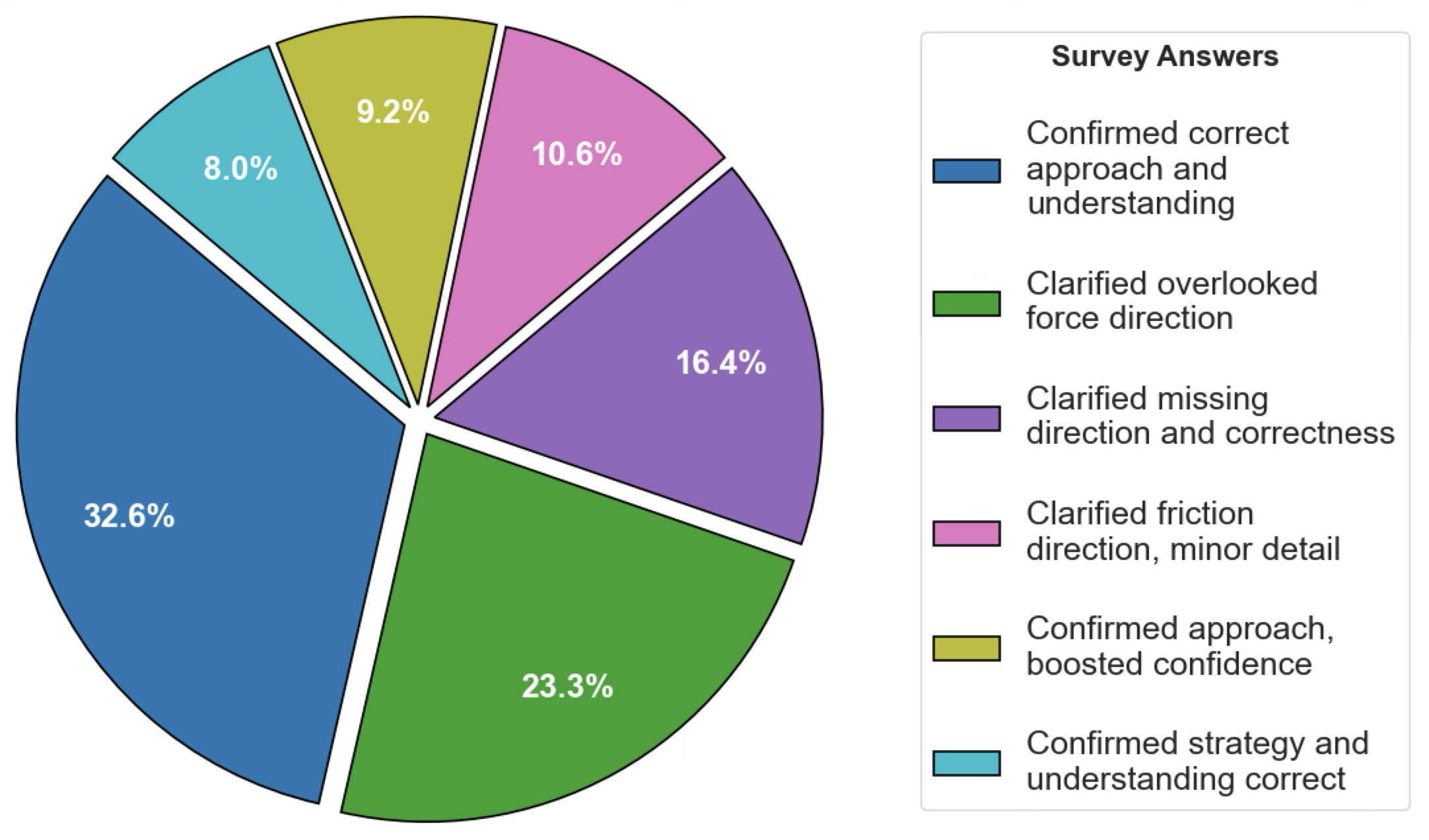}
    \caption{Survey results regarding why LLM's feedback was helpful. Aside from confirmation, students indicate that LLM was useful in clarifying overlooked / missing aspects in the essays}
    \label{fig:survey-positive-reasons}
\end{figure}
To measure the effectiveness of LLM-generated feedback, we compare student performance on the same quiz question over four consecutive semesters. As shown in Figure \ref{fig:performance-improvement}, more than half of the class in previous semesters made errors. In contrast, the error rate drops to less than 10\% with the help of JiT LLM intervention. 

Notably, the feedback remains highly effective despite sub-optimal performance of LLMs in predicting error types. We attribute this to domain-knowledge grounding, which incorporated expert-generated feedback that addresses potentially missing or incorrect concepts regardless of the predicted error class.

After students solve the quiz with LLM feedback, we ask whether the feedback was helpful and why. Approximately 77.59\% of the students reported the feedback was helpful; their qualitative responses are clustered using HDBSCAN, and shown in Figure \ref{fig:survey-positive-reasons}. 

There are two major themes emerged from the clusters. Around half of the students valued the feedback for concept verification and confirmation, while the other half stated that the LLMs clarified overlooked or missing items in their strategy essays. This illustrates the adaptiveness of JiT LLM feedback; for students with a correct understanding, the LLM provides essential validation, whereas for those prone to errors, it guides them rethink or pay more attention to specific concepts.

\subsection{Learning through Conversations}

\begin{table}[t]\footnotesize
\centering
\begin{tabular}{lc}
\toprule
\textbf{Metric} & \textbf{Value} \\
\midrule
Total instances & 1,042 \\
Conversational instances  & 209 (20.05\%) \\
Mean \# conv. turns (std) & 2.77 (1.52) \\
Min/Max \# conv. turns & 2 / 14 \\
Conv. turn dist. skewness & 3.57  \\
\midrule
\textbf{\% of Model Prediction as \texttt{Correct}} & \\
Turn 1 (Initial essay) & 42.79\% \\
Last turn (Final essay) & 71.64\% \\
\bottomrule
\end{tabular}
\caption{Statistics of conversational instances. The skewness of the number of conversation turn distributions is computed by Fisher-Pearson coefficient of skewness where the value greater than 2 is normally considered extremely skewed.}
\label{tab:conv_summary}
\end{table}

One of the emergent behaviors of students in interacting with feedback-generating LLM is that they have conversations with the LLMs to improve their understanding. This observation is impressive considering that students are not directed or incentivized to iteratively chat with the LLMs. Students are notified as solving the quiz with LLMs will give them a bonus credit regardless of the correctness of their answers. Nevertheless, there are around 20\% of the students who voluntarily spent more time interacting with the LLM so that they could improve their learning experience. Basic statistics of the conversational instances are reported in Table \ref{tab:conv_summary}, and more detailed statistics are reported in Appendix \ref{sec:appendix-conversation}.

Figure \ref{fig:conv-sample} shows an example of conversation between a student and LLM. As the student receives LLM's feedback that tackles missing or incorrect points in the strategy essay, they add these components to their strategy essay. By continuing conversations, students incrementally add missing points and gain more comprehensive understandings. Table \ref{tab:conv_summary} illustrates such instances: while only 43\% of students' initial essays were predicted as \texttt{correct}, the portion boosted to 72\% after having conversations.

\begin{figure*}[t]
    \centering
    \includegraphics[width=0.9\textwidth]{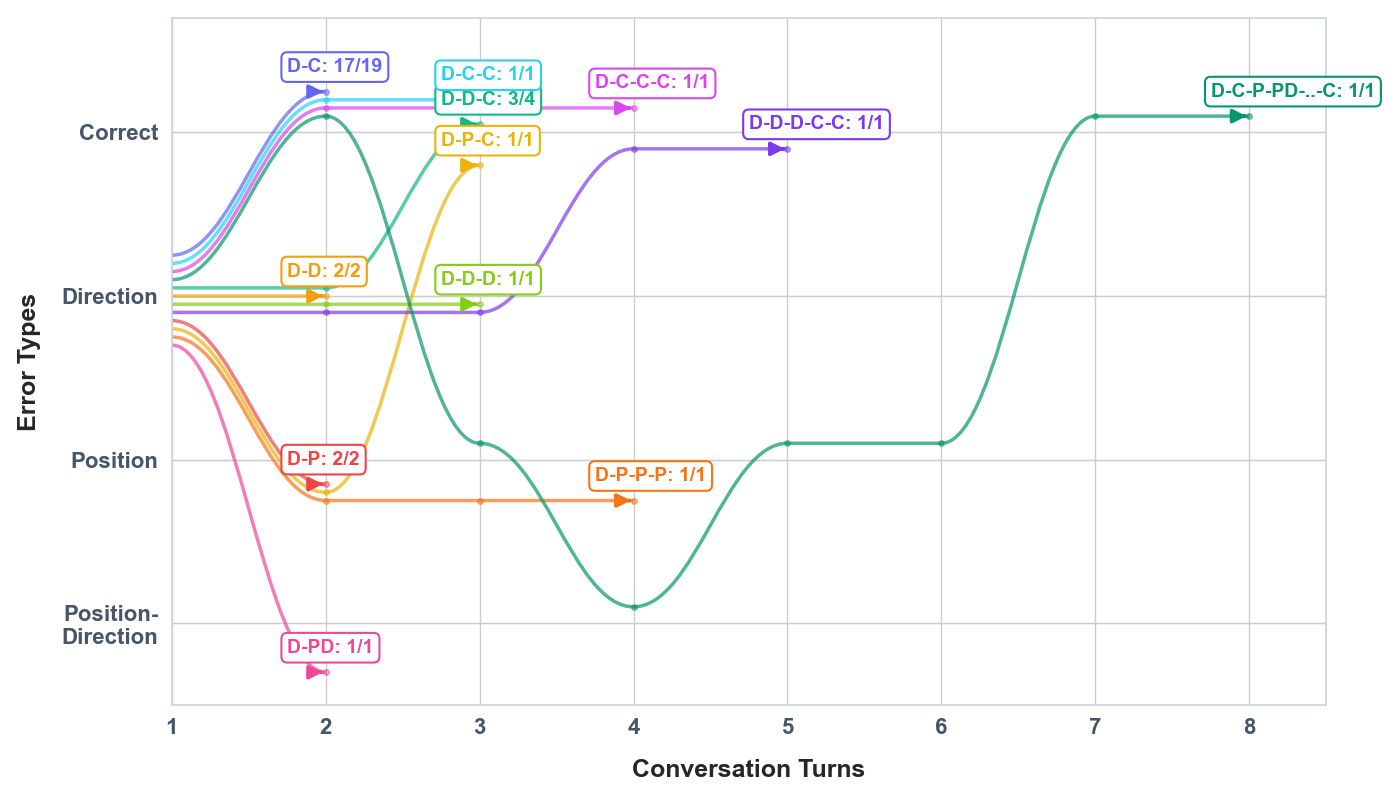}
    \caption{    \textbf{Learning Trajectories} for students starting with \texttt{Direction} error. Each trace represents a sequence of essay error types as predicted by the feedback-LLM at each conversation turn. Terminal markers are labeled with the specific Path Sequence and the Path Outcome. For instance, `D-C: 17/19' indicates that out of 19 students whose essays the LLM classified as `\texttt{Direction} error' and then `\texttt{Correct}', 17 got correct answers in the end.}
    \label{fig:discussion-traces}
\end{figure*}

We visualize learning trajectories by analyzing the sequence of conversation turns with the LLM. This involves tracking the intermediate strategy essays and their corresponding predicted error types throughout the interaction. Figure \ref{fig:discussion-traces} illustrates how students modify their strategy essays and how the predicted error types change while they have conversations with LLMs. The figure represents a subset where students' strategy essays are initially classified as having a \texttt{direction} error. 

As conversation proceeds, students keep modifying their essays by applying LLM's advice, and there exists multiple learning trajectories based on how they improve their essays. In the end, approximately 80\% of the students corrected their essays and the LLM classified them as `\texttt{correct}'. Other students, even though their final version of essays are not classified as `\texttt{correct}', they solved the quiz correctly. We assume that this is the case where students skip the final validation step after having their perspectives corrected. Overall, 91.43\% of the students (32 out of 35) solved the quiz correctly by iteratively getting feedback from LLM and developing their understandings.

We conducted a further analysis to validate the hypothesis that students who engage more actively with the LLM feedback improve their strategy essays better. We defined ``activeness'' through two metrics: reflective latency (the time in seconds between conversation turns) and degree of revision (the change in word count between consecutive essay versions). Specifically, we posit that spending more time refining an essay or making more substantial text changes indicates a more active adaptation of the feedback. 

\begin{figure}[t]
    \centering
    \includegraphics[width=\columnwidth]{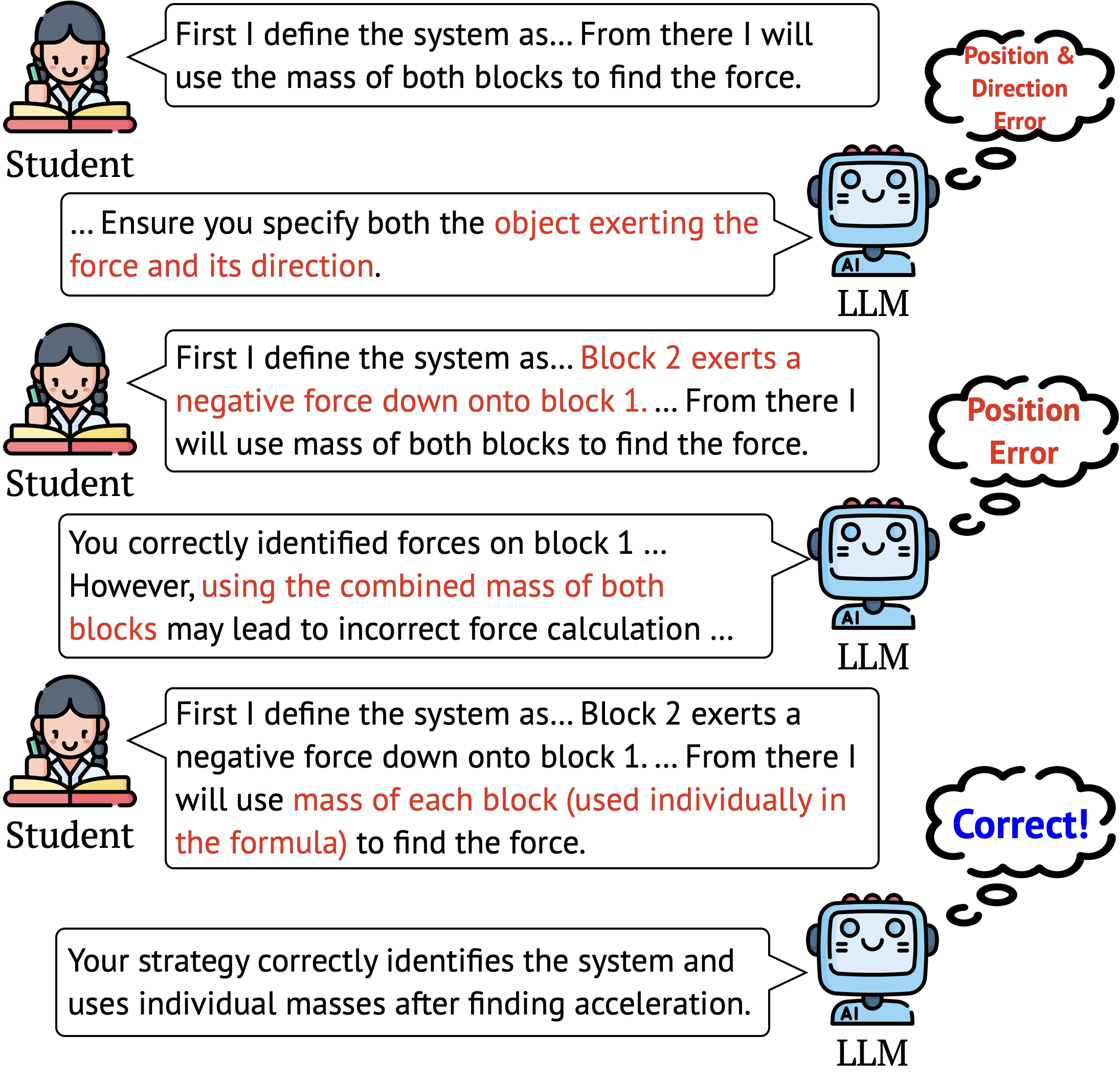}
    \caption{    \textbf{Example Conversation} of a student modifying their strategy essays with the help of LLM. LLM generates feedback based on its error type prediction, and informs different aspects as the student fixes errors one by one.}
    \label{fig:conv-sample}
\end{figure}

\begin{figure}[H]
    \centering
    \includegraphics[width=\columnwidth]{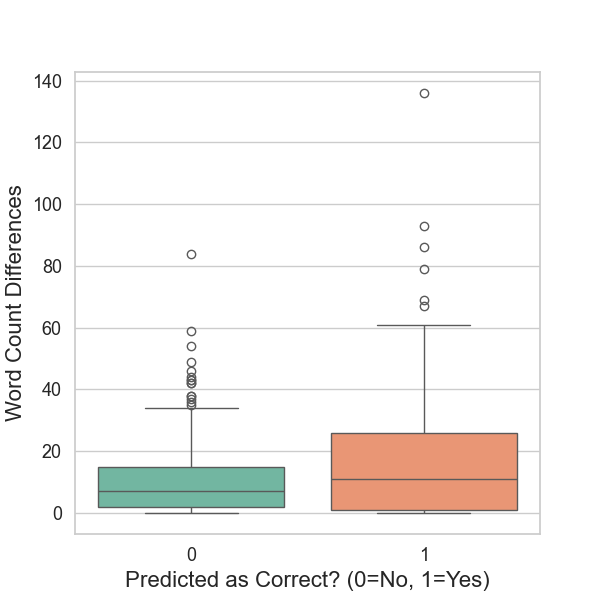}
    \caption{\textbf{Correlation between word count differences and essay correctness.} We compute the correlation between the word count differences in strategy essays between each conversation utterances and the correctness of essay. When the students make more changes to their essays, they are more likely getting their essays correct.}
    \label{fig:wordcntdiff}
\end{figure}
To evaluate these hypotheses, we compute the Pearson correlation coefficients. The results indicate a negligible relationship between time spent and essay correctness ($r = 0.06$, $p = 0.27$). In contrast, the correlation between word count difference and essay correctness is statistically significant ($r = 0.18$, $p < 0.001$). Figure \ref{fig:wordcntdiff} illustrates how word count differences are distributed differently for the essays that are classified as \texttt{correct} or not. These findings suggest that simply spending more time on the interface is insufficient; rather, the degree of revision is the primary factor of success. This implies that while the LLM provides the necessary feedback, the likelihood of achieving a correct understanding depends heavily on the student's willingness to actively reformulate their reasoning.

\section{Conclusion}

In this paper, we explore how LLMs can be utilized efficiently in higher education through Just-in-Time (JiT) adaptive interventions. By analyzing strategy essays from students, our framework identifies potential errors and provides adaptive feedback grounded in domain knowledge. Experimental results show that our framework significantly improves the overall student performance by providing validation and guiding re-evaluation of incorrect concepts. Qualitative analysis further demonstrates that students engage in conversations with LLMs to improve their understanding, highlighting the potential for LLMs to serve as sophisticated pedagogical tools.

\section*{Limitations}
One of the limitations of this work is the sub-optimal accuracy of the error type classification. Predicting error types solely from strategy essays is a non-trivial task even for humans. Students often omit important details, whether due to oversight or lack of conceptual understanding. LLM feedback falls into a catastrophic failing mode when the LLM incorrectly classifies an essay as `\texttt{correct}' and does not provide necessary interventions as a result of that. While we mitigate this risk by considering second most-likely label, the reliability and robustness of the framework can be improved with better classification. In future iterations, we could improve it by using relevant metadata (e.g. student's level of knowledge, tone and vocabulary used in the essay, etc.), or by enforcing a more structured essay format to ensure all necessary concepts are articulated.

Another limitation is the small number of JiT interventions, as the framework was applied to only a single quiz during the semester. Although we observe significant performance improvements with the LLM feedback, a broader deployment across multiple topics and quiz types is necessary to validate these findings. Expanding the scope would allow for a more granular analysis of which specific subject matters or problem types benefit most from LLM-based interventions, providing deeper insights in how LLM feedback should be designed.

Our student preference survey shows that while the LLM is highly effective for novice-level feedback, it struggles to generate feedback that is more suitable to students with advanced knowledge; students with over-the-average performance still preferred LLM feedback that is targeted to novice-level over the advanced-level. This opens a new direction for designing LLM feedback in future studies to provide minimal interventions considering the knowledge level of students.

From a technical point of view, approximately 4\% of the students experienced server failures due to the high volume of synchronous API calls, mainly due to the size of the course being very large ($N > 1,000$). To make the framework more scalable, future implementations need to address rate-limiting and asynchronous processing. Furthermore, the current LLM lacks a memory function that help efficient conversations; the current version processes updated strategy essays in isolation without the context from previous versions. In future work, we aim to incorporate student-specific memory and conversation history that reduces the number of turns required for a student to reach a correct understanding.

\section*{Ethics Consideration}
This research was declared `Exempt' under IRB protocol number IRB-2023-391. The participant pool consisted of undergraduate students enrolled in a large-scale introductory Physics course. All personally identifiable information, including student names and university IDs, was removed from the data before being processed by the LLMs. Student essays used for qualitative analysis in this paper have been further anonymized, truncated, and paraphrased. 

Our framework contains risks when generating feedback to students, primarily because of hallucinations and incorrect classification of error types. To mitigate these risks, the LLM was grounded with expert-annotated domain knowledge as well as its secondary predictions. Furthermore, we prioritized equity of treatment and algorithmic fairness throughout the study. The LLM was prompted using a standardized, invariant template for all participants, ensuring that the feedback was controlled solely by the student's essay rather than any external variables such as student's personal information or prior performance levels. By maintaining a uniform prompting strategy, we ensured that no specific group of students was prioritized or discriminated against, providing consistent pedagogical support to the entire students.

We disclose that LLMs are used for checking grammatical errors and exploring paraphrases. All core research contributions including motivation, experiment designs, and results analyses were developed by the authors.

\section*{Acknowledgments}

This research is supported in part by U.S. National Science Foundation grant 2300645. Opinions expressed are of the authors and not of the Foundation.
\bibliography{custom}

@inproceedings{taneja2024jill,
  title={Jill watson: A virtual teaching assistant powered by chatgpt},
  author={Taneja, Karan and Maiti, Pratyusha and Kakar, Sandeep and Guruprasad, Pranav and Rao, Sanjeev and Goel, Ashok K},
  booktitle={International Conference on Artificial Intelligence in Education},
  pages={324--337},
  year={2024},
  organization={Springer}
}

@inproceedings{kweon2025large,
  title={A large-scale real-world evaluation of an llm-based virtual teaching assistant},
  author={Kweon, Sunjun and Nam, Sooyohn and Lim, Hyunseung and Hong, Hwajung and Choi, Edward},
  booktitle={Proceedings of the 63rd Annual Meeting of the Association for Computational Linguistics (Volume 6: Industry Track)},
  pages={850--864},
  year={2025}
}

@article{kestin2025ai,
  title={AI tutoring outperforms in-class active learning: An RCT introducing a novel research-based design in an authentic educational setting},
  author={Kestin, Greg and Miller, Kelly and Klales, Anna and Milbourne, Timothy and Ponti, Gregorio},
  journal={Scientific Reports},
  volume={15},
  number={1},
  pages={17458},
  year={2025},
  publisher={Nature Publishing Group UK London}
}

@inproceedings{zhang2025simulating,
  title={Simulating classroom education with llm-empowered agents},
  author={Zhang, Zheyuan and Zhang-Li, Daniel and Yu, Jifan and Gong, Linlu and Zhou, Jinchang and Hao, Zhanxin and Jiang, Jianxiao and Cao, Jie and Liu, Huiqin and Liu, Zhiyuan and others},
  booktitle={Proceedings of the 2025 Conference of the Nations of the Americas Chapter of the Association for Computational Linguistics: Human Language Technologies (Volume 1: Long Papers)},
  pages={10364--10379},
  year={2025}
}

@article{kalyuga2007expertise,
  title={Expertise reversal effect and its implications for learner-tailored instruction},
  author={Kalyuga, Slava},
  journal={Educational psychology review},
  volume={19},
  number={4},
  pages={509--539},
  year={2007},
  publisher={Springer}
}

@inproceedings{albacete2019impact,
  title={The impact of student model updates on contingent scaffolding in a natural-language tutoring system},
  author={Albacete, Patricia and Jordan, Pamela and Katz, Sandra and Chounta, Irene-Angelica and McLaren, Bruce M},
  booktitle={International conference on artificial intelligence in education},
  pages={37--47},
  year={2019},
  organization={Springer}
}

@inproceedings{walker2012noticing,
  title={Noticing relevant feedback improves learning in an intelligent tutoring system for peer tutoring},
  author={Walker, Erin and Rummel, Nikol and Walker, Sean and Koedinger, Kenneth R},
  booktitle={International Conference on Intelligent Tutoring Systems},
  pages={222--232},
  year={2012},
  organization={Springer}
}

@article{wei2022chain,
  title={Chain-of-thought prompting elicits reasoning in large language models},
  author={Wei, Jason and Wang, Xuezhi and Schuurmans, Dale and Bosma, Maarten and Xia, Fei and Chi, Ed and Le, Quoc V and Zhou, Denny and others},
  journal={Advances in neural information processing systems},
  volume={35},
  pages={24824--24837},
  year={2022}
}

@inproceedings{pang2025physics,
  title={Physics reasoner: Knowledge-augmented reasoning for solving physics problems with large language models},
  author={Pang, Xinyu and Hong, Ruixin and Zhou, Zhanke and Lv, Fangrui and Yang, Xinwei and Liang, Zhilong and Han, Bo and Zhang, Changshui},
  booktitle={Proceedings of the 31st International Conference on Computational Linguistics},
  pages={11274--11289},
  year={2025}
}

@inproceedings{arora2023have,
  title={Have llms advanced enough? a challenging problem solving benchmark for large language models},
  author={Arora, Daman and Singh, Himanshu and others},
  booktitle={Proceedings of the 2023 Conference on Empirical Methods in Natural Language Processing},
  pages={7527--7543},
  year={2023}
}

@article{singh2016multiple,
  title={Multiple-choice test of energy and momentum concepts},
  author={Singh, Chandralekha and Rosengrant, David},
  Journal = {Am. J. Phys.}, Volume = {71}, Number = {6}, Pages = {607-617}, Month = {June}, Year = {2003}
}

@article{brooks2019matrix,
  title={A matrix of feedback for learning},
  author={Brooks, Cameron and Carroll, Annemaree and Gillies, Robyn M and Hattie, John},
  journal={Australian Journal of Teacher Education (Online)},
  volume={44},
  number={4},
  pages={14--32},
  year={2019}
}

@article{burgess2020feedback,
  title={Feedback in the clinical setting},
  author={Burgess, Annette and van Diggele, Christie and Roberts, Chris and Mellis, Craig},
  journal={BMC medical education},
  volume={20},
  number={Suppl 2},
  pages={460},
  year={2020},
  publisher={Springer}
}

@book{hattie2018visible,
  title={Visible learning: feedback},
  author={Hattie, John and Clarke, Shirley},
  year={2018},
  publisher={Routledge}
}

@article{henderson2021usefulness,
  title={The usefulness of feedback},
  author={Henderson, Michael and Ryan, Tracii and Boud, David and Dawson, Phillip and Phillips, Michael and Molloy, Elizabeth and Mahoney, Paige},
  journal={Active Learning in Higher Education},
  volume={22},
  number={3},
  pages={229--243},
  year={2021},
  publisher={SAGE Publications Sage UK: London, England}
}

@article{kluger1996effects,
  title={The effects of feedback interventions on performance: a historical review, a meta-analysis, and a preliminary feedback intervention theory.},
  author={Kluger, Avraham N and DeNisi, Angelo},
  journal={Psychological bulletin},
  volume={119},
  number={2},
  pages={254},
  year={1996},
  publisher={American Psychological Association}
}

@article{wisniewski2020power,
  title={The power of feedback revisited: A meta-analysis of educational feedback research},
  author={Wisniewski, Benedikt and Zierer, Klaus and Hattie, John},
  journal={Frontiers in psychology},
  volume={10},
  pages={487662},
  year={2020},
  publisher={Frontiers}
}

@article{furnborough2009adult,
  title={Adult beginner distance language learner perceptions and use of assignment feedback},
  author={Furnborough, Concha and Truman, Mike},
  journal={Distance Education},
  volume={30},
  number={3},
  pages={399--418},
  year={2009},
  publisher={Taylor \& Francis}
}

@article{hattie2007power,
  title={The power of feedback},
  author={Hattie, John and Timperley, Helen},
  journal={Review of educational research},
  volume={77},
  number={1},
  pages={81--112},
  year={2007},
  publisher={Sage Publications Sage CA: Thousand Oaks, CA}
}

@article{orlando2016comparison,
  title={A comparison of text, voice, and screencasting feedback to online students},
  author={Orlando, John},
  journal={American Journal of Distance Education},
  volume={30},
  number={3},
  pages={156--166},
  year={2016},
  publisher={Taylor \& Francis}
}

@article{ortiz2005college,
  title={College students’ perceptions of quality in distance education: The importance of communication},
  author={Ortiz-Rodr{\'\i}guez, Madeline and Telg, Ricky W and Irani, Tracy and Roberts, T Grady and Rhoades, Emily},
  journal={Quarterly Review of Distance Education},
  volume={6},
  number={2},
  pages={97--105},
  year={2005},
  publisher={Emerald Publishing Limited}
}

@article{pitt2017now,
  title={‘Now that’s the feedback I want!’Students’ reactions to feedback on graded work and what they do with it},
  author={Pitt, Edd and Norton, Lin},
  journal={Assessment \& Evaluation in Higher Education},
  volume={42},
  number={4},
  pages={499--516},
  year={2017},
  publisher={Taylor \& Francis}
}

@article{price2010feedback,
  title={Feedback: all that effort, but what is the effect?},
  author={Price, Margaret and Handley, Karen and Millar, Jill and O'donovan, Berry},
  journal={Assessment \& Evaluation in Higher Education},
  volume={35},
  number={3},
  pages={277--289},
  year={2010},
  publisher={Taylor \& Francis}
}

@inproceedings{ryan2016written,
  title={Written feedback doesn’t make sense’: Enhancing assessment feedback using technologies},
  author={Ryan, Tracii and Henderson, Michael and Phillips, Michael},
  booktitle={International conference of the Australian Association for Research in Education},
  pages={1--11},
  year={2016}
}

@article{winstone2017d,
  title={‘It'd be useful, but I wouldn't use it’: barriers to university students’ feedback seeking and recipience},
  author={Winstone, Naomi E and Nash, Robert A and Rowntree, James and Parker, Michael},
  journal={Studies in Higher Education},
  volume={42},
  number={11},
  pages={2026--2041},
  year={2017},
  publisher={Taylor \& Francis}
}

@inproceedings{wolsey2008efficacy,
  title={Efficacy of instructor feedback on written work in an online program},
  author={Wolsey, Thomas},
  booktitle={International Journal on E-learning},
  volume={7},
  number={2},
  pages={311--329},
  year={2008},
  organization={Association for the Advancement of Computing in Education (AACE)}
}

@inproceedings{docktor2010conceptual,
  title={A conceptual approach to physics problem solving},
  author={Docktor, Jennifer L and Strand, Natalie E and Mestre, Jos{\'e} P and Ross, Brian H},
  booktitle={AIP Conference Proceedings},
  volume={1289},
  number={1},
  pages={137--140},
  year={2010},
  organization={American Institute of Physics}
}

@article{dufresne1992constraining,
  title={Constraining novices to perform expertlike problem analyses: Effects on schema acquisition},
  author={Dufresne, Robert J and Gerace, William J and Hardiman, Pamela Thibodeau and Mestre, Jose P},
  journal={The Journal of the Learning Sciences},
  volume={2},
  number={3},
  pages={307--331},
  year={1992},
  publisher={Taylor \& Francis}
}

@article{leonard1996using,
  title={Using qualitative problem-solving strategies to highlight the role of conceptual knowledge in solving problems},
  author={Leonard, William J and Dufresne, Robert J and Mestre, Jose P},
  journal={American Journal of Physics},
  volume={64},
  number={12},
  pages={1495--1503},
  year={1996},
  publisher={American Association of Physics Teachers}
}

@article{mestre1993promoting,
  title={Promoting skilled problem-solving behavior among beginning physics students},
  author={Mestre, Jose P and Dufresne, Robert J and Gerace, William J and Hardiman, Pamela T and Touger, Jerold S},
  journal={Journal of research in science teaching},
  volume={30},
  number={3},
  pages={303--317},
  year={1993},
  publisher={Wiley Online Library}
}

@inproceedings{rebello2019using,
  title={Using a Hybrid of Argumentation and Problem Solving Prompts to Facilitate Undergraduates' Problem Solving Performance and Confidence},
  author={Rebello, Carina M and Piedrahita Uruena, Yuri},
  booktitle={The 13th Conference of the European Science Education Research Association (ESERA)},
  year={2019}
}

@article{tuminaro2007elements,
  title={Elements of a cognitive model of physics problem solving: Epistemic games},
  author={Tuminaro, Jonathan and Redish, Edward F},
  journal={Physical Review Special Topics—Physics Education Research},
  volume={3},
  number={2},
  pages={020101},
  year={2007},
  publisher={APS}
}

@article{chi1981categorization,
  title={Categorization and representation of physics problems by experts and novices},
  author={Chi, Michelene TH and Feltovich, Paul J and Glaser, Robert},
  journal={Cognitive science},
  volume={5},
  number={2},
  pages={121--152},
  year={1981},
  publisher={Elsevier}
}

@article{foster2024impact,
  title={The impact of formative assessment on student learning outcomes: A meta-analytical review},
  author={Foster, H},
  journal={Academy of Educational Leadership Journal},
  volume={28},
  number={S1},
  pages={1--3},
  year={2024}
}

@inproceedings{phung2024automating,
  title={Automating human tutor-style programming feedback: Leveraging gpt-4 tutor model for hint generation and gpt-3.5 student model for hint validation},
  author={Phung, Tung and P{\u{a}}durean, Victor-Alexandru and Singh, Anjali and Brooks, Christopher and Cambronero, Jos{\'e} and Gulwani, Sumit and Singla, Adish and Soares, Gustavo},
  booktitle={Proceedings of the 14th learning analytics and knowledge conference},
  pages={12--23},
  year={2024}
}

@inproceedings{jia2024llm,
  title={LLM-generated feedback in real classes and beyond: Perspectives from students and instructors},
  author={Jia, Qinjin and Cui, Jialin and Du, Haoze and Rashid, Parvez and Xi, Ruijie and Li, Ruochi and Gehringer, Edward},
  booktitle={Proceedings of the 17th international conference on educational data mining},
  pages={862--867},
  year={2024}
}

@inproceedings{dai2023can,
  title={Can large language models provide feedback to students? A case study on ChatGPT},
  author={Dai, Wei and Lin, Jionghao and Jin, Hua and Li, Tongguang and Tsai, Yi-Shan and Ga{\v{s}}evi{\'c}, Dragan and Chen, Guanliang},
  booktitle={2023 IEEE international conference on advanced learning technologies (ICALT)},
  pages={323--325},
  year={2023},
  organization={IEEE}
}

@inproceedings{Hashmi2025, Author = "Syed Furqan Abbas Hashmi and N. Sanjay Rebello", Title = {Analyzing Undergraduate Problem-Solving in Physics Through Interaction With an AI Chatbot}, BookTitle = {Physics Education Research Conference 2025}, Pages = {184-189}, Address = {Washington, DC}, Series = {PER Conference}, Month = {August 6-7}, Year = {2025} }

\appendix

\section{Feedback Preference Survey}
\label{sec:appendix-preference-survey}
\begin{figure}[htbp]
    \centering
    \includegraphics[width=\columnwidth]{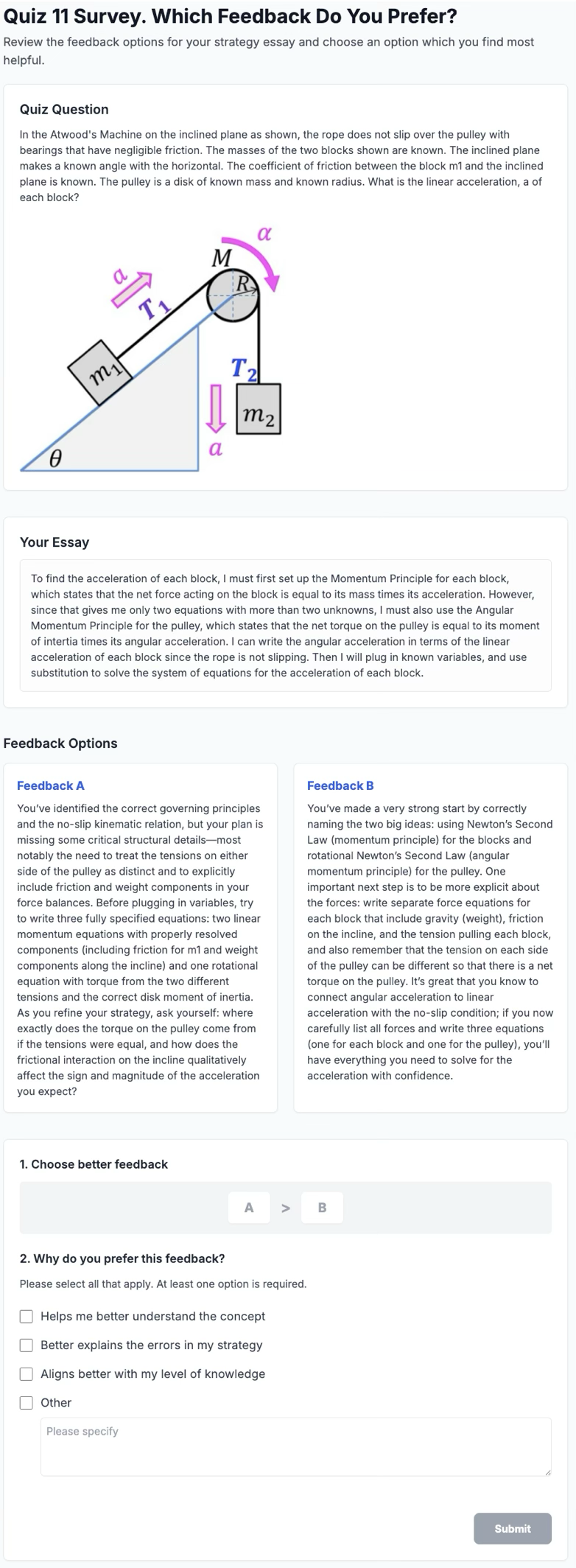}
    \caption{Survey webpage. Students answer two questions; choosing better feedback and the reason why.}
    \label{fig:survey-webapp}
\end{figure}

\begin{figure*}[ht]\footnotesize
    \begin{promptbox}{Prompt Sample: Generating Post-hoc Feedback for Preference Survey}
\texttt{You are an expert College Teaching Assistant and Pedagogical Specialist. Your goal is to provide feedback on a student's "Strategy Essay" regarding a quiz problem. First, analyze the student's written strategy to infer their Knowledge Level based on the complexity and accuracy of their writing. Then, regardless of their actual level, generate two distinct versions of feedback (one adapted for a "Novice" and one for an "Advanced" student) to provide options for instruction. Generate each feedback in 3 sentences.}

\vspace{1em}
\textbf{Input Data Provided:}
\begin{enumerate}
    \item The Quiz Problem: The specific question the student attempted to solve.
    \item Student's Strategy Essay: The student's written explanation of how they approached the problem. 
    \item Expert's Strategy Essay: A sample strategy essay written by an expert, that can serve as a rubrics for evaluating student's essay 
\end{enumerate}

\textbf{Instructions for Adaptation:}
\begin{itemize}
    \item If Knowledge Level is ``NOVICE'': 
    \begin{itemize}
        \item Tone: Highly encouraging, supportive, and patient. Use the ``Sandwich Method'' (Praise -> Gentle Correction -> Praise). 
        \item Vocabulary: Avoid jargon where possible. If technical terms are used, briefly define them or use analogies. 
        \item Focus: Focus on identifying misconceptions and solidifying the fundamental concept. 
        \item Scaffolding: Break down the next steps into small, manageable pieces. Guide them toward the correct starting point. 
        \item Goal: To build confidence and ensure they understand the basic mechanism of the problem. 
    \end{itemize}
    \item If Knowledge Level is "ADVANCED": 
    \begin{itemize}
        \item Tone: Professional, peer-to-peer, challenging, and concise. 
        \item Vocabulary: Use precise, high-level academic/technical terminology appropriate for the field. 
        \item Focus: Focus on optimization, edge cases, and connecting this problem to broader course concepts. Point out subtle logic flaws or inefficiencies. 
        \item Scaffolding: Do not hand-hold. Ask Socratic questions that force the student to re-evaluate their own logic. 
        \item Goal: To refine their reasoning, improve efficiency, and deepen metacognition. 
    \end{itemize}
\end{itemize}

\textbf{Feedback Structure (Output Format) Please output the response in the following json format:}\\
\{ ``Essay\_Evaluation'': ``1-2 sentences summarizing their approach based on the expert sample'', \\
``Inferred\_Level'': ``Either Novice or Advanced'', \\
``Feedback'': \{\\
``Novice'': ``The core feedback paragraph. Apply the Instructions for Adaptation logic for NOVICE here strictly.'', \\
``Advanced'': ``The core feedback paragraph. Apply the Instructions for Adaptation logic for ADVANCED here strictly.'', \\
\}
\}

\vspace{1em}
[INPUT DATA]\\
\textbf{The Quiz Problem:} {\texttt{quiz\_description\_input}}\\
\textbf{Student's Strategy Essay:} {\texttt{student\_strategy\_input}}\\
\textbf{Expert's Strategy Essay:} {\texttt{expert\_strategy\_with\_rubrics}}
    \end{promptbox}
    \caption{The full system prompt used for generating post-hoc feedback}
    \label{fig:ad-hoc-prompt}
\end{figure*}

We survey student's LLM feedback preference in the Fall 2025 semester for two quizzes. Students are instructed that they will get extra credits for participating in the survey, and the number of participants for each iteration was $1,074$ and $1,160$. The two options for feedback is generated with the prompt described in Figure \ref{fig:ad-hoc-prompt}; it first evaluates the student's strategy essays using expert-annotated rubrics, then generate two different versions of feedback using gpt-5.1, one for novice and the other for advanced.

Once the feedback generation is completed, we hosted a separate survey webpage from the student's standard Learning Management System as shown in Figure \ref{fig:survey-webapp}. The two options are randomly swapped to minimize any potential biases. Students are instructed to read the quiz problem and essay they authored to remind themselves of the context, and choose one of the feedback texts they prefer with a set of pre-defined reasons (students can choose multiple reasons). For both surveys, the most selected reason was ``Helps me better understand the concept'', followed by ``Better explains the errors in my strategy'' and ``Aligns better with my level of knowledge''.

\section{Error Type Prediction Model Details}
\label{sec:appendix-model-accuracy}
For fine-tuned classifiers, we randomly split the data into 80-10-10 ratio for train-validation-test sets. We use DeBERTa-v3-small model with learning rate of 2e-5 and batch size of 16. The model is fine-tuned and tested on a single NVIDA A100 GPU with 80GB memory, takes around 10 minutes to fine-tune 5 epochs.

For LLM-based inference, we used the deepseek-r1:70B checkpoint, hosted on-premise at the authors' institution, to make sure the student's essay data is not shared outside. 

All fine-tuning and inferences are performed with 2 trials and the average score was reported with the confidence interval in Table \ref{tab:model_results}.

\section{JiT Feedback Implementation}
\label{sec:appendix-live-feedback}

\begin{figure*}[ht]\footnotesize
\begin{promptbox}{Prompt Sample: Generating Just-in-Time Feedback for Quiz Problem Solving}
\texttt{You are an expert Physics Education Researcher and Cognitive Scientist. Your goal is to analyze student essays describing their strategy for solving a problem. You must determine if the student's proposed strategy will lead to the Correct Answer or a specific type of Misconception.}

\vspace{1em}
\textbf{TASK DESCRIPTION:} I will provide you with a physics problem statement and a ``Strategy Essay'' written by a student.
\begin{enumerate}
    \item Analyze the student's essay. Look for specific keywords or logical steps.
    \item Identify if they are applying correct principles/concepts or falling into misconceptions.
    \item Output your final reasoning and label as strictly valid JSON.
\end{enumerate}

\textbf{Label Categories \& Definitions}
\begin{itemize}
    \item `correct': The reasoning and final answer are physically sound.
    \item `direction': The student makes an error where their answer is in the exact opposite direction of the correct vector OR The student does not specifically state the direction they want to apply.
    \item `position': The error arises from calculating forces or properties on a different object than the one asked OR The student does not specifically state the particular object they want to calculate for.
    \item `position-direction': Both a `position' error and a `direction' error occur simultaneously in the student's response OR The student does not specifically state the direction nor the object they want to consider.
\end{itemize}

\textbf{Few-Shot Examples}

[Examples of student's sample strategy essay, its error type, and feedback generated by human experts]

\textbf{TEST CASE}
\{quiz\_problem\}\\
\{STUDENT\_STRATEGY\_ESSAY\}

\textbf{INSTRUCTIONS:} Analyze the ``Student Essay'' above and provide the JSON output.\\
\textbf{OUTPUT FORMAT:}\\
\{
    ``classification'': Choose one from [correct, direction, position, position-direction],\\
    ``confidence'': How confident the classification is (scale of 1 to 5, 5 being most confident),\\
    ``secondary\_classification'': What would be the second most likely label?\\
    ``feedback'': Feedback message (in 50 words) to students based on their essay. If their essay is classified to make a certain type of mistake, the feedback should highlight that aspect. If it is classified as correct, the feedback can be general and notify some common mistakes that can be made. If the secondary classification is different from classification, take both classes into consideration.\\
\}

\textbf{Constraint}
\begin{itemize}
    \item Output ONLY the json.
    \item Do NOT provide reasoning, explanations, or introductory text.
\end{itemize}

\end{promptbox}
\caption{The full system prompt used for generating JiT feedback}
\label{fig:JiT-prompt}
\end{figure*}

\begin{figure*}[htbp]
    \centering
    \includegraphics[width=\linewidth]{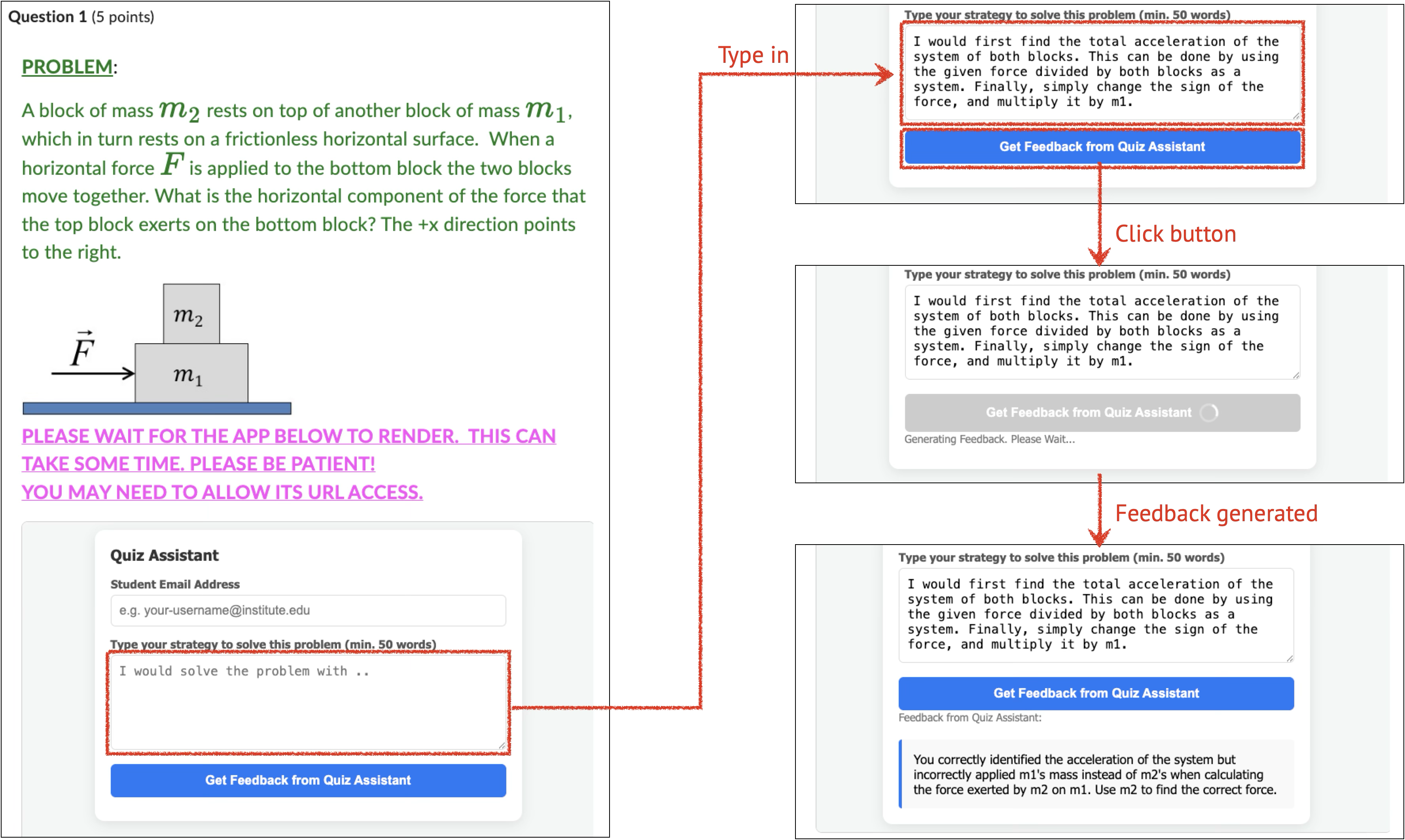}
    \caption{Overall flow of JiT feedback. The feedback snippet is placed below the quiz question. Students type in their strategy essays, click a button, and gets a feedback after a few seconds.}
    \label{fig:feedback-prompt}
\end{figure*}

When the LLM is deployed for generating JiT feedback, we host a web app using Flask API, which is later embedded to the Learning Management System that students use for the course. The web app is a simple text box where students put their strategy essays and clicks a button to get the feedback from the LLM. In the back-end, we use pre-defined few-shot examples with Chain-of-Thought reasoning so that the model considers primary and secondary class of the strategy essay. The example of an actual prompt and the quiz page students have accessed is shown in Figure \ref{fig:JiT-prompt} and Figure \ref{fig:feedback-prompt}.

\section{Conversation Analysis}
\label{sec:appendix-conversation}

\begin{table}[ht]\footnotesize
\centering
\begin{tabular}{lcccc}
\toprule
\textbf{From \textbackslash To} & \textbf{Correct} & \textbf{Position} & \textbf{Direction} & \textbf{Pos-Dir} \\
\midrule
\textbf{Correct} & \textbf{0.6797} & 0.1328 & 0.1250 & 0.0625 \\
\textbf{Position} & 0.3491 & \textbf{0.3585} & 0.1415 & 0.1509 \\
\textbf{Direction} & \textbf{0.5278} & 0.1250 & 0.2917 & 0.0556 \\
\textbf{Pos-Dir} & \textbf{0.3750} & 0.2969 & 0.0781 & 0.2500 \\
\bottomrule
\end{tabular}
\caption{Transition probabilities between error types across conversation turns}
\label{tab:transition_matrix}
\end{table}
In this section, we report additional analyses on conversational instances. Table \ref{tab:transition_matrix} shows the transition probabilities between conversation turns. When the student's strategy essay at the $i$-th conversation turn is predicted as \texttt{correct}, it is still going to be classified as the same label with the $67.97\%$ chances, but there exists slightly more than $30\%$ chances the essay is classified as one of the error types. While majority of the strategy essays with \texttt{direction} and \texttt{position-direction} errors are likely developed into \texttt{correct} strategy, essays with \texttt{position} error is more likely going to stay at the same error type. This implies that LLM's ability to guide students avoid errors differ based on the types.




\end{document}